\documentclass[11pt]{article}
\usepackage{wnut2020}
\usepackage{multicol}
\usepackage{multirow}
\usepackage{xcolor}
\usepackage{amsfonts}
\usepackage{amsmath}
\usepackage{times}
\usepackage{url}
\usepackage{hyperref}

\usepackage{cleveref}

\usepackage{caption}
\usepackage{subcaption}

\usepackage{latexsym}
\usepackage{graphicx}

\date{}

\begin{document}
\maketitle

\begin{abstract}
In this paper, we describe our approach in the shared task: \textit{COVID-19 event extraction from Twitter}. The objective of this task is to extract answers from COVID-related tweets to a set of predefined slot-filling questions. Our approach treats the event extraction task as a question answering task by leveraging the transformer-based T5 text-to-text model.
  
According to the official evaluation scores returned, namely F1, our submitted run achieves competitive performance compared to other participating runs (Top 3). However, we argue that this evaluation may underestimate the actual performance of runs based on text-generation. Although some such runs may answer the slot questions well, they may not be an exact string match for the gold standard answers. To measure the extent of this underestimation, we adopt a simple exact-answer transformation method aiming at converting the well-answered predictions to exactly-matched predictions. The results show that after this transformation our run overall reaches the same level of performance as the best participating run and state-of-the-art F1 scores in three of five COVID-related events. Our code is publicly available to aid reproducibility\footnote{\url{https://github.com/wangcongcong123/ttt/tree/master/covid_event}}.
 
\end{abstract}
\section{Introduction}

Since the outbreak of COVID-19, a wide variety of research has been conducted to mine insights or gain rapid access to information in relation to the crisis. For example, the TREC-COVID challenge seeks advanced techniques for finding useful information from the CORD-19 corpus comprising hundreds of thousands of COVID-related academic articles~\cite{wang2020cord19,Kirk2020}. Related research focuses on mining valuable information from social media~\cite{muller2020covid,dimitrov2020tweetscov19}. This is largely inspired by the fact that social media motivates people to share information or express their opinions quickly. However, this user-generated content is usually noisy and enormous during crises, and thus needs to be condensed and filtered before further processing and analysis. 

\begin{figure}
    \centering
    \includegraphics[scale=0.8]{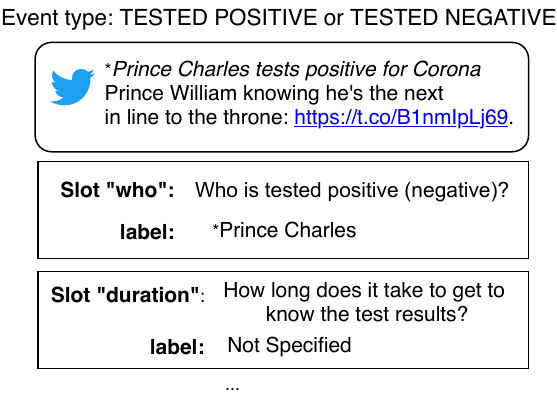}
    \caption{An example of the event extraction task: this shows a tweet represents a \texttt{TESTED POSITIVE} or \texttt{TESTED NEGATIVE} event. The objective is to extract answers to the slot questions concerning the event.}
    \label{fig:task-desc}
\end{figure}

Motivated by this, the ACL Workshop on Noisy User-generated Text 2020 proposed a shared task (W-NUT Task-3) of \textit{extracting COVID-19 related events from Twitter}~\cite{zong2020extracting}. The major objective of this task is to seek computational linguistic techniques for extracting text spans from a corpus of raw tweets to answer a set of predefined slot questions. The corpus used in the task can be described in two parts. First, the corpus consists of approximately 7,500 tweets categorised into five broad event types: (1) \texttt{TESTED POSITIVE}, (2) \texttt{TESTED NEGATIVE}, (3) \texttt{CAN NOT TEST}, (4) \texttt{DEATH} and (5) \texttt{CURE AND PREVENTION}. As the names indicate, these tweets are a summary of people’s primary concerns about COVID-19 that they likely post about on social media. Secondly, for the tweets in each event, a set of questions or slot-filling
types are defined to help gather more fine-grained information about the tweets. The human annotations, (i.e.,  ground truths) of the corpus are simply the answers to the predefined questions. Figure~\ref{fig:task-desc} illustrates an example from this corpus. This example represents a TESTED POSITIVE or TESTED NEGATIVE event, and the associated slot-filling questions relate to the ``who'' and the ``duration'' slots, asking who has tested positive or negative and how long it takes to know the test result. The annotation process in part 2 is conducted by annotators who select answers from a drop-down list of candidate choices\footnote{The choices are automatically-extracted text spans obtained through a Twitter tagging tool~\cite{ritter2011named} or predefined choices such as ``not specified'', ``yes'', ``no'', etc. For more details, see ~\cite{zong2020extracting}.}. This explains why the label for slot ``who'' in Figure~\ref{fig:task-desc} has the symbol $*$ at the beginning.

Using this corpus, we leverage a text-to-text model for generating answers to the slot questions. Given that recent years have witnessed the success of transformers~\cite{vaswani2017attention,BERT2018,raffel2019exploring} in natural language processing (NLP) via transfer learning, we adopt the T5~\cite{raffel2019exploring} model architecture and its pre-trained weights. The method is straightforward but effective, easy to adapt to domain-similar tasks, and efficient in data input without needing additional pre-processing and part-of-speech features. In this paper, we report its performance as compared to other participating runs. Our run achieves competitive performance as indicated by F1. However, evaluation is based on exact matches, and thus a prediction is deemed a true positive only when it is a character-by-character match with the ground truth from the candidate answers. Since our method treats the task as a text-generation-based question answering task instead of categorising labels from a fixed list of candidate labels, it potentially generates well-answered predictions that do not exactly match the ground truth answers. Having observed some of these mismatches in our experiment we subsequently transformed these to exact-matched answers using a simple approach based on Levenshtein string edit distance~\cite{levenshtein1966binary}. After this transformation, our best run reaches the same level of F1 performance as the best participating run.


\begin{figure*}[!h]
    \centering
    \includegraphics{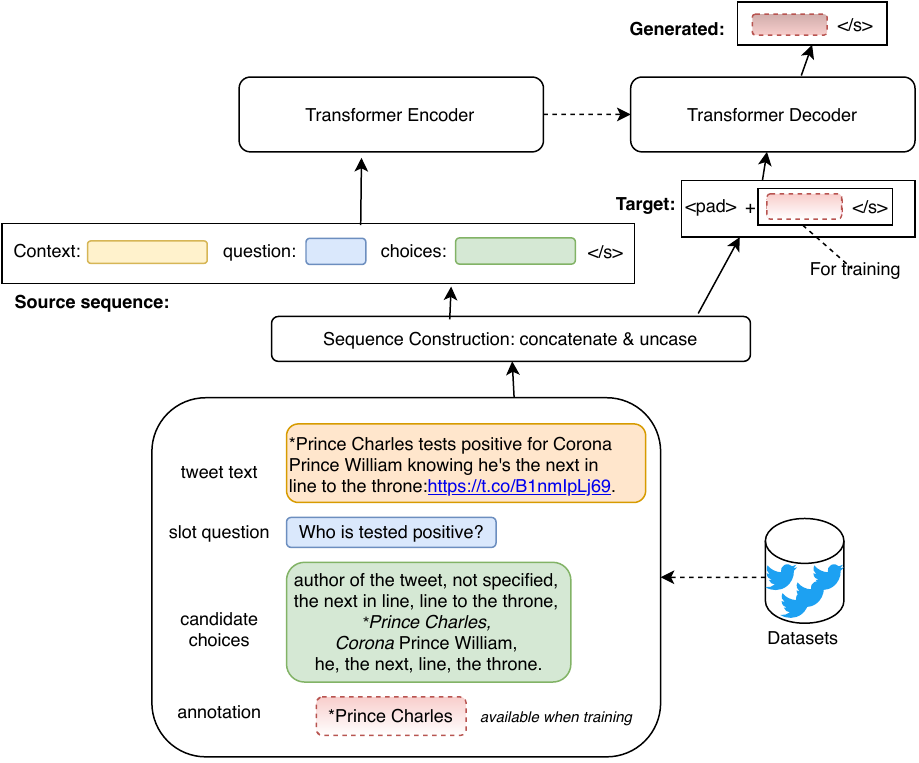}
    \caption{The system architecture of our approach}
    \label{fig:sys-arch}
\end{figure*}
\section{Related Work}

The literature has seen much work on processing crisis-related messages on social media. For example, the TREC Incident Streams~\cite{McCreadie2020} track is a research initiative for seeking computational linguistic techniques for finding actionable information from Twitter during crises. With similar motivations to this initiative, many techniques have been applied for crisis messages categorisation and analysis in recent years. For instance, \newcite{miyazaki2019label} apply label embedding for crisis tweet classification adopting a bi-directional LSTM  model. \newcite{congcong2020cls} leverage contextual ELMO embeddings~\cite{peters2018deep} and data augmentation. CrisisBERT is proposed for crisis event detection~\cite{liu2020crisisbert} through fine-tuning the pre-trained BERT~\cite{BERT2018}.

More recently, the COVID-19 pandemic has motivated research on insights mining, information extraction and classification on social media. Since the pandemic is characterised by being long-lived, unlike more short-term crises such as earthquakes or shootings, the messages on social media are studied to gain a better understanding of the pandemic from multiple perspectives. For example, TweetsCOV19~\cite{dimitrov2020tweetscov19} is a corpus of semantically annotated tweets about COVID-19 that can potentially be used to explore public opinion and perception on the pandemic. \newcite{muller2020covid} introduce COVID-Twitter-BERT, which pre-trains a BERT model on hundreds of millions of COVID-related tweets and then fine-tunes the model to downstream tasks including vaccine sentiment analysis and vaccine stance classification.

Although most recent work involves fine-tuning transformer-based models for COVID-related short message processing, we have chosen to take an alternative approach for this task. Inspired by the text-to-text T5 model, our approach formulates the W-NUT Task 3 as a question-answering problem. The principal idea is that its unified source and target structure can be easily adapted to other domain-similar tasks such as informativeness classification or vaccine sentiment for COVID-related tweets. Regarding fine-tuning T5 for COVID-related language tasks, probably the most relevant work is that \newcite{tang2020rapidly} fine-tuned T5 for COVID-related question answering from scientific articles. However, they constructed the source sequence by concatenating a document and a query, and the target sequence is a generated token indicating the document's relevance to the query. Our approach is characterised by being capable of unifying various text classification tasks to a general question answering task. In our approach, the source sequence takes the raw text as the context, a classification description as the question and the classification's labels as the candidate choices. The target sequence is simply the generated answers that are exact text spans from the source sequence, indicating which label(s) the text belongs to.

\section{Method}
In this section, we detail our approach in W-NUT Task-3. We firstly describe how a general transformer encoder-decoder model work in a sequence-to-sequence way and then introduce our approach adapting it to the event extraction task. Basically, given a source sequence $X\colon\{x_1,x_2,...,x_n\}$, the transformer encoder uses it as the input to output its contextualised encoding sequence $\overline{X}\colon\{\overline{x}_1,\overline{x}_2,...,\overline{x}_n\}$. The encoding process can be represented by a mapping function $f$ with learnable parameters $\theta_{\text{e}}$, as follows.

$$ f_{\theta_{\text{e}}}: \mathbf{X}_{1:n} \to \mathbf{\overline{X}}_{1:n}. $$

The encoded source sequence then is used as the input to the transformer decoder that defines the conditional probability distribution of a target sequence $Y\colon\{y_1,y_2,...,y_m\}$  given $\overline{X}_{1:n}$:

$$ p_{\theta_{d}}(\mathbf{Y}_{1:m} | \mathbf{\overline{X}}_{1:n}) = \prod_{i=1}^{m} p_{\theta_{d}}(\mathbf{y}_i | \mathbf{Y}_{0: i-1}, \mathbf{\overline{X}}_{1:n}). $$

Where $p$ with learnable parameters $\theta_{\text{d}}$ is the function to learn the conditional probability distribution by the decoder. The target token $y_i$ is generated conditional on both the source context $\overline{X}_{1:n}$ and its previous $i-1$ already-generated tokens in an auto-regressive way. The powerful part of the transformer-based decoder is that it applies a single directional attention mechanism, i.e., masked multi-head attention as proposed by~\newcite{vaswani2017attention} to learn the relations between the target token and its previous tokens. Meanwhile, it applies a bi-directional attention mechanism, i.e., multi-head attention, to learn the relations between the target token and the source context, which is also used in the encoder to learn the relations (self-attentions) between the source tokens.

Inspired by the feature of target generation given a source in a transformer encoder-decoder model, we leverage it for the shared task. Figure~\ref{fig:sys-arch} presents the system architecture of our approach. The core component of the system is the sequence construction, which converts each tweet in the dataset into a source sequence and target sequence that well fits to train the transformer encoder-decoder model in a text-to-text fashion. As introduced, each tweet in the dataset is annotated in two parts, 1), labels indicating the event types, 2), answers to the slot questions. Our approach uses both parts of the annotations to construct the source and target sequences as follows:

\begin{itemize}
    \item \textbf{Source}: This sequence is constructed from the raw tweets through mapping their major four attributes, i.e., as outlined in Figure~\ref{fig:sys-arch}, the tweet text is mapped to ``context'', event type question (part 1) or slot question (part 2)\footnote{The event type questions are manually constructed: taking \texttt{DEATH} as an example: \textit{Does this tweet report death from coronavirus?}} to ``question'', and the candidate choices to ``choices''. The leads to the final source sequence concatenating all these fields in the form: ``context: \{tweet text\} question: \{slot/event question\} choices: \{candidates\}".
    
    \item \textbf{Target}: The target sequence needs both part 1 and part 2 annotations in training. They are simply mapped from the event labels for part 1 (``yes'' or ``no'' indicating if the tweet falls into the event as specified in the question field of the source sequence) and annotation answers for part 2. Where the annotations are not available in inference, the source sequence and the target sequence with the start decoder token \texttt{<pad>} are fed as input to the model for generating the answers directly.
\end{itemize}

Although the aforementioned only details the sequence construction process using the event extraction corpus as an example, it follows similar steps when adapting to similar tasks. For example, for a binary informativeness classification task, the question field can be replaced by something like ``Is the tweet informative of COVID-19?'' and choices can be ``yes'' or ``no'' implying \texttt{INFORMATIVENESS} or \texttt{UNINFORMATIVENESS}. We leave this generalisation capability to future work and here continue to focus on the event extraction task.

After the source and target sequences are constructed, the next step is to train a transformer encoder-decoder model. As evidenced in the literature, sequence-to-sequence transformers such as BART~\cite{lewis2019bart} or T5~\cite{raffel2019exploring} has demonstrated great success in various downstream language tasks via transfer learning. Since T5 has different pre-trained weights available where in particular its small version is easy to handle in the pilot study, we set up the encoder-decoder model following the T5~\cite{raffel2019exploring} architecture and train it on this task via fine-tuning its pre-trained weights.


\section{Experiments}
This section describes the details of experimental process and results.

\subsection{Data Preparation}

Because the original dataset is released with only Tweets IDs, some tweets had become invalid prior to our retrieval time and consequently 7,149 valid tweets were used in our experiments. Since each training tweet is assigned one event type label and multiple slot annotations, there a total of 34,464 examples used in our experiments after source and target sequence construction. Among these, we sample 10\% as the validation set and the rest as the training set.

\subsection{Training}

\begin{table}[!h]
\small
\centering
\begin{tabular}{|l|l|}
\hline
Batch   size       & 16                  \\ \hline
Epoch              & 12                  \\ \hline
Learning rate (LR) & 5e-05            \\ \hline
LR scheduler       & Warmup linear decay \\ \hline
Warmup ratio       & 0.10                \\ \hline
Optimizer        & Adam\\ \hline
\end{tabular}
\caption{Hyper-parameters of model training}
\label{tab:hyper-parameters}
\end{table}

In our experiments, we chose \texttt{t5-small}, \texttt{t5-base}, and \texttt{t5-large} for fine-tuning. Table~\ref{tab:hyper-parameters} shows the hyper-parameter configuration in our experiments. The batch size is selected from the options of $\{8,16,32\}$, based on evaluation on the validation set and the other hyper-parameters are determined with reference to the literature in a similar domain~\cite{liu2020crisisbert}. We fine-tune each model with 12 epochs as we observe no further improvements after this. The maximum source and target sequence lengths were set to be 473 and 78 respectively since no examples exceeded these. The training was accelerated by a TPU3-8, taking around 4 hours to complete the \texttt{t5-large} fine-tuning. After the model is fine-tuned, we use greedy decoding during inference. This is because the generated texts are usually short answers in our problem and thus top-k or top-p sampling~\cite{holtzman2019curious} would give similar results as greedy decoding.

\begin{table}[!h]
\small
   \begin{subtable}[h]{0.45\textwidth}
        \centering
      
\begin{tabular}{|l|r|r|r|l|}
\hline
                 & \multicolumn{1}{l|}{F1} & \multicolumn{1}{l|}{P} & \multicolumn{1}{l|}{R} & Params    \\ \hline
small-2          & 0.5308                  & 0.4308                 & 0.6913                 & 1.0x  \\ \hline
base-2           & 0.6225                  & 0.5449                 & 0.7258                 & 3.7x  \\ \hline
\textbf{large-2} & 0.6392                  & 0.5800                 & 0.7118                 & 12.8x \\ \hline
\end{tabular}
        \caption{Evaluation results of different sizes of models where \textbf{large-2} is the officially submitted \textbf{run-2} and Params refers to the model's parameters relative to \texttt{t5-small} that has around 60M parameters.}
        \label{tab:our-results-sizes}
     \end{subtable}
      \hfill
     \vspace{6pt}

    \begin{subtable}[h]{0.45\textwidth}
        \centering
   \begin{tabular}{|l|l|l|l|}
\hline
team name     & F1     & P      & R      \\ \hline
winners        & 0.6598$\star$ & 0.7272 & 0.6039 \\ \hline
HLTRI          & 0.6476 & 0.7532$\star$ & 0.5679 \\ \hline
\textbf{Our (run-2)}        & 0.6392 & 0.5800 & 0.7118$\star$ \\ \hline
VUB            & 0.6160 & 0.6875 & 0.5580 \\ \hline
UPennHLP       & 0.5237 & 0.6754 & 0.4277 \\ \hline
Test\_Positive & 0.5114 & 0.5377 & 0.4875 \\ \hline
\end{tabular}
       \caption{Evaluation results of submitted runs ranked by F1. \textbf{Our (run-2)} is named \textbf{UCD\_CS} officially.}
       \label{tab:submitted-results}
    \end{subtable}
    
    \hfill
    \vspace{6pt}

    \begin{subtable}[h]{0.45\textwidth}
        \centering
 \begin{tabular}{|l|l|l|l|}

\hline
Ours           & F1     & P      & R      \\ \hline
run-1          & 0.6429$\star$ & 0.5815 & 0.7188~$\star$ \\ \hline
\textbf{run-2} & 0.6392 & 0.5800 & 0.7118 \\ \hline
run-3          & 0.6367 & 0.5920$\star$ & 0.6887 \\ \hline
\end{tabular}
        \caption{Evaluation results of our \texttt{t5-large} runs at different epochs.}
        \label{tab:our-results}
     \end{subtable}
     
      \hfill
     \vspace{6pt}
     
     \begin{subtable}[h]{0.45\textwidth}
        \centering
\begin{tabular}{|l|l|l|l|}
\hline
Ours           & F1     & P      & R      \\ \hline
post-run-1          & 0.6571$\star$ & 0.5956 & 0.7327$\star$ \\ \hline
\textbf{post-run-2} & 0.6517 & 0.5921 & 0.7247 \\ \hline
post-run-3          & 0.6495 & 0.6050$\star$ & 0.7012 \\ \hline
\end{tabular}
       \caption{Evaluation results of our \texttt{t5-large} runs after post-processing.}
       \label{tab:post-preprocessed-results}
    \end{subtable}

     \caption{F1, precision (P) and recall (R) scores averaged over the five COVID-19 events where $\star$ refers to the highest in each column.}
     \label{tab:eval-results}
     
\end{table}

\subsection{Results and Discussion}

We saved the last three checkpoints of the final fine-tuned T5 models ready for evaluation at epochs 10, 11, and 12 (denoted as \textbf{run-3}, \textbf{run-2}, and \textbf{run-1} for \texttt{t5-large} respectively). The evaluation was conducted on a test set of 2,500 tweets (500 per event) for part-2 only. Since this shared task only allows one submission from each team and we found trivial difference between \textbf{run-2} and  \textbf{run-1} based on the validation evaluation, we submitted \textbf{run-2} for the official evaluation. Table~\ref{tab:submitted-results} reports the F1, recall (R) and precision (P) scores of the submitted runs. Additionally, we put the evaluation results of all our three \texttt{t5-large} runs in Table~\ref{tab:our-results} as well as of different sizes of T5 models in Table~\ref{tab:our-results-sizes} for reference\footnote{The small-2 and base-2 corresponds to \texttt{t5-small} and \texttt{t5-base} respectively that are fine-tuned with the same experimental setup as \textbf{run-2}} (discussed in Section~\ref{subsec:post-processing}).
\begin{table*}[tb]
\small
\centering
\begin{tabular}{c|p{3.5cm}|p{10cm}}
\hline\hline

\multirow{5}{*}{Example 1} & Tweet text             & only precautionary steps can protect and prevent us from corona virus. sanitation, mask,  alertness, yoga and healthy food.\\
& Slot question (where)  & \textbf{what is the cure for coronavirus mentioned by the author of the tweet?}      
\\ 
& Our raw prediction   & \textcolor{orange}{mask,  alertness, yoga and healthy food} \\& Post-processed prediction & \textcolor{blue}{virus. sanitation, mask,  alertness, yoga and healthy food} \\ 
& Ground truth & \textcolor{violet}{virus. sanitation, mask,  alertness, yoga and healthy food} \\ \hline\hline

\multirow{5}{*}{Example 2} & Tweet text             & @bbhuttozardari what is your sindh government doing? stop playing politics at this time and take measures to prevent spread of corona. provide healthcare facilities. sometimes doing something is better than barking tweets.\\
& Slot question (what)  & \textbf{what is the cure for coronavirus mentioned by the author of the tweet?}
\\ 
& Our raw prediction   & \textcolor{orange}{provide healthcare facilities} \\& Post-processed prediction & \textcolor{blue}{. provide healthcare facilities} \\ 
& Ground truth & \textcolor{violet}{. provide healthcare facilities} \\ \hline\hline

\multirow{5}{*}{Example 3} & Tweet text             & @realdonaldtrump they’re drinking bleach to cure covid-19?.\\
& Slot question (what)  &\textbf{ what is the cure for coronavirus mentioned by the author of the tweet?} 
\\ 
& Our raw prediction  & \textcolor{orange}{bleach} \\& Post-processed prediction & \textcolor{blue}{@realdonaldtrump they’re drinking bleach} \\ 
& Ground truth & \textcolor{violet}{@realdonaldtrump they’re drinking bleach} \\ \hline\hline

\multirow{5}{*}{Example 4} & Tweet text             & @joshua4congress my sister is a vet with an active-duty husband and a 6wk old baby. she has a fever and symptoms but can't get tested bc she lives in a military base in texas. they require exposure to a confirmed positive. she had to give birth without family the day after our grandma died.\\
& Slot question (where)  & \textbf{where is the can’t-be-tested situation reported?}     
\\ 
& Our raw prediction   & \textcolor{orange}{a military base in texas} \\& Post-processed prediction & \textcolor{blue}{a military base} \\
& Ground truth & \textcolor{violet}{texas} \\
\hline\hline

\multirow{5}{*}{Example 5} & Tweet text             & so in a few weeks time 4 year olds will be expected to be back in school but can't get tested. doesn't make sense. \#covid19 \#dailybriefings\\
& Slot question (who)  & \textbf{who can not get a test?}     
\\ 
& Our raw prediction   & \textcolor{orange}{4 year olds} \\& Post-processed prediction & \textcolor{blue}{year olds} \\
& Ground truth & \textcolor{violet}{a few weeks time 4 year olds} \\
\hline\hline
\end{tabular}
\caption{Examples of mismatches (uncased), i.e., predictions that are self-evidently correct answers, but do not exactly match the ground truths in \textcolor{violet}{violet text}. The \textcolor{orange}{orange text} refers to the original submitted predictions generated by our approach and the \textcolor{blue}{blue text} refers to the transformed predictions from the raw predictions based on their edit distances to candidate answers.}
\end{table*}

\subsubsection{Performance}

First, Table~\ref{tab:our-results-sizes} indicates that the larger the model is, the better overall performance it can achieve. Interestingly, as compared to small-2, large-2's advantage over base-2 seems not significant (F1: 0.6392 versus 0.6225) given its 12.8x larger size versus base-2's 3.7x larger size. Hence, our base-2 can achieve competitive performance with a decent number of parameters. Here we continue to focus on the large runs since we want to explore the limit of our approach's performance.

Table~\ref{tab:submitted-results} presents the evaluation results of our \texttt{t5-large} based runs at different epochs. It shows that our submitted run (i.e., run-2) achieved competitive performance as compared to other participating runs (F1: our 0.6392 versus the highest 0.6598). Particularly, our run exhibited a significant advantage in recall over other runs (0.7118 versus the second-highest of 0.6039). However, this advantage is combined with a trade-off in terms of precision. This reveals that our run was somewhat ``active'' at finding the answers to the slot questions. This is consistent with its performance at event type level as well. Table~\ref{tab:results-event-type} shows the evaluation  results  of  our run-2  at  event  type  level (we will discuss post-run-2 in Section~\ref{subsec:post-processing}). We find that our run achieves the best recall in every event type but quite behind in precision, especially for \texttt{can\_not\_test}, \texttt{death} and \texttt{cure}. To further analyse the cause of low precision, we rethink the metrics used to evaluate runs with a similar approach to ours.

\subsubsection{Post Processing}\label{subsec:post-processing}

\begin{table*}[t]
    \centering
    \small
\begin{tabular}{|l|r|r|r|r|r|r|r|r|r|}
\hline
               & \multicolumn{3}{c|}{F1}                                                                  & \multicolumn{3}{c|}{P}                                                                   & \multicolumn{3}{c|}{R}                                                                   \\ \hline
               & \multicolumn{1}{c|}{best} & \multicolumn{1}{c|}{run-2} & \multicolumn{1}{l|}{post-run-2} & \multicolumn{1}{c|}{best} & \multicolumn{1}{c|}{run-2} & \multicolumn{1}{l|}{post-run-2} & \multicolumn{1}{c|}{best} & \multicolumn{1}{c|}{run-2} & \multicolumn{1}{l|}{post-run-2} \\ \hline
positive       & 0.6973                    & 0.6778                     & \textbf{0.6989}                          & 0.8569                    & 0.7380                     & 0.7620                          & 0.6267                    & 0.6267                     & \textbf{0.6454}                          \\ \hline
negative       & 0.7030                    & 0.7030                     & \textbf{0.7047}                          & 0.7107                    & 0.6873                     & 0.6890                          & 0.7194                    & 0.7194                     & \textbf{0.7212}                          \\ \hline
can\_not\_test & 0.6523                    & 0.5660                     & 0.5667                          & 0.6863                    & 0.4646                     & 0.4656                          & 0.7240                    & 0.7240                     & 0.7240                          \\ \hline
death          & 0.6942                    & 0.6048                     & 0.6191                          & 0.7240                    & 0.4917                     & 0.5041                          & 0.7855                    & 0.7855                     & \textbf{0.8020}                          \\ \hline
cure           & 0.6205                    & 0.6078                     & \textbf{0.6236}                       & 0.8405                    & 0.4961                     & 0.6236                          & 0.7843                    & 0.7843                     & \textbf{0.8028}                          \\ \hline
\end{tabular}
    \caption{The evaluation results of our run-2 and post-run-2 at event type level where best represents the highest score across all participating runs. These in bold stand for new state-of-the-art scores in W-NUT task-3 after applying \textbf{TransM}.}
    \label{tab:results-event-type}
\end{table*}
It is worth noting that, the part 2 annotations were originally labeled by providing a fixed list of candidate choices. For example, the candidate choices in Figure~\ref{fig:sys-arch} are either pre-defined (author of the tweet, not specified) or text spans extracted from the raw tweet via the named entity tagging tool~\cite{ritter2011named}. This makes it easy to treat the task as a classification-based slot filling task, i.e, binary categorising if a candidate choice answers a given slot type~\cite{zong2020extracting}. In this sense, the F1, P and R are good metrics for evaluating the performance of a system in this task. However, our approach is based on answer-generation and thus it is likely to generate some correct predictions that are not exact character matches. In situations where a mismatch occurs this is doubly penalised by the metrics. The prediction made by our system will be considered to be a false positive, because it does not match a label from the gold standard. Simultaneously the presence of a gold standard label that our system does not return results in a false negative being counted also. We argue this can result in the metrics underestimating the effective performance of such systems as it has an adverse effect on precision, recall and F1 score.

Table~\ref{tab:post-preprocessed-results} presents a list of examples from \textbf{run-1} where this occurs. In these examples, the raw generated predictions are good answers to the corresponding slot questions, although they do not exactly match the ground truth from the candidate options. In particular, in example 2, the answer ``provide healthcare facilities'' is taken as a false positive as it misses the dot character at the beginning. In examples 3 and 5, our raw prediction is arguable a better answer than the ground truth. Example 1 is especially interesting in that our raw prediction omits one important word (``sanitation'') whereas the ground truth includes an additional word that is not part of the most appropriate question answer (``virus''). To alleviate this effect, we apply a simple post-processing method to transform the raw predictions to these that can  best be exactly matched with the ground truth labels. The transformation method (which we name ``TransM'') is described as follows.

\textbf{TransM}: Given the raw prediction $r$ for an example $x$ and its candidate choices $c: [c_1,c_2,..c_i.,c_n]$ where $n$ is the number of candidates, $r$ is converted to the final prediction $t$ that is the one selected from $c$ with the shortest edit distance to $p$. The edit distance is simply calculated by the Levenshtein~\cite{levenshtein1966binary} distance divided by the length of $c_i$.

After this post-processing, we can now see from Table~\ref{tab:post-preprocessed-results} that most of the post-processed predictions have been transformed to exactly-matched predictions. As a result, we re-evaluate our T5 large runs with the transformed predictions and present the results in Table~\ref{tab:post-preprocessed-results} and report run-2 performance at event type level after applying TransM, i.e., post-run-2 in Table~\ref{tab:results-event-type}. First, Table~\ref{tab:post-preprocessed-results} shows that our three runs can achieve higher scores overall as compared to the original predictions-based runs. Also, our best run post-run-1 can actually reach the same level of performance as the best participating run (F1: 0.6571 versus 0.6598). Referring to Table~\ref{tab:results-event-type}, it presents that not only does our post-run-2 achieve noticeable improvements across F1, P, and R in almost every event as compared to run-2 but it hits new state-of-the-art F1 scores in the \texttt{positive}, \texttt{negative} and \texttt{cure} events. Notably, for \texttt{positive} and \texttt{cure}, our run-2 was around 2 points behind the best participating runs in F1 but it reached the new state-of-the-art F1 scores after applying TransM (0.6989 and 0.6236). This reveals that the exact-matched metrics were exerting stronger underestimation in measuring our system's performance for answering tested-positive and cure-and-prevention related questions than other types of questions. 

Although the post-processing alleviates the underestimation of performance, our run is not the best in precision. Arguably, the last two rows (example 4-5) in Table~\ref{tab:post-preprocessed-results} are good examples showing that in some cases that our raw predictions contain the ground truths but are taken as false positives despite the post-processing: i.e., TransM helps alleviate the underestimation but not completely eliminate it. We present a complete list of unmatched predictions generated by run-2 or post-run-2 to enable the community to examine the system's performance openly and critically\footnote{\url{https://github.com/wangcongcong123/ttt/tree/master/covid_event}}.



\section{Conclusion}
This paper presents our text-to-text based approach at W-NUT 2020 shared task 3. We show that the principal idea behind the approach is adaptability to other domain-similar tasks such as informativeness classification of COVID-19 tweets. We expect to conduct more work on this adaptability in the future. It is even more interesting to test the idea in zero-shot learning. For example, how well it performs if transferring the model that is trained on the event extraction corpus to do inference in the informativeness task directly without further training. In addition, we empirically present that our system is effective, achieving competitive performance and arguably the state-of-the-art F1 scores in three of five COVID-events in the shared task. Despite the effectiveness, one concern of our approach is the model size. Our best performed model is fine-tuned using the large version of T5 with around 770M parameters~\cite{li2020train}. This makes it important to compress the model efficiently in the future.

\section*{Acknowledgments}
We would like to thank Google’s TensorFlow Research Cloud (TFRC) team who provided TPUs credits to support this research.

\bibliographystyle{acl_natbib} 
\bibliography{wnut2020}


\appendix


\end{document}